\documentclass[runningheads,a4paper]{llncs}

\usepackage{graphicx} 
\usepackage{subfigure} 

\usepackage{algorithm}
\usepackage{algorithmic}

\usepackage{color}
\usepackage{amsmath}
\usepackage{amsbsy}
\usepackage{amssymb}

\usepackage{eucal}
\usepackage{url}
\usepackage{array}
\usepackage{enumerate}

\newtheorem{mydef}{Definition}

\newcommand{\sX}{{\mathcal{X}}}
\newcommand{\sY}{{\mathcal{Y}}}
\newcommand{\sZ}{{\mathcal{Z}}}

\newcommand{\bB}{{\mathbf{B}}}
\newcommand{\bI}{{\mathbf{I}}}

\newcommand{\bR}{{\mathbf{R}}}
\newcommand{\bC}{{\mathbf{C}}}
\newcommand{\bX}{{\mathbf{X}}}

\newcommand{\bSigma}{{\mathbf{\Sigma}}}

\newcommand{\bx}{{\boldsymbol{x}}}
\newcommand{\by}{{\boldsymbol{y}}}

\newcommand{\be}{{\boldsymbol{e}}}
\newcommand{\br}{{\boldsymbol{r}}}

\newcommand{\sfa}{{\sffamily{\bf{(a)}}}}
\newcommand{\sfb}{{\sffamily{\bf{(b)}}}}

\newcommand{\exm}{\mu}

\newcommand{\cov}{\mbox{cov}}

\newcommand{\nsamp}{m}

\newcommand{\Perp}{\perp \! \! \! \perp}
\newcommand{\notPerp}{\hspace{1.5mm} / \hspace{-4.5mm} \Perp}

\begin{document}

\mainmatter  

\title{Estimating a Causal Order among Groups of Variables in Linear Models}

\titlerunning{Estimating a Causal Order among Groups of Variables in Linear Models}

%
%
\author{Doris Entner \and Patrik O. Hoyer}
\authorrunning{Doris Entner, Patrik O. Hoyer}

\institute{HIIT \& Department of Computer Science, University of Helsinki, Finland}

%
%

\toctitle{Estimating a Causal Order among Groups of Variables in Linear Models}
\tocauthor{Doris Entner, Patrik O. Hoyer}
\maketitle

\begin{abstract}
The machine learning community has recently devoted much attention to the problem of inferring causal relationships from statistical data. Most of this work has focused on uncovering connections among scalar random variables. We generalize existing methods to apply to collections of multi-dimensional random \emph{vectors}, focusing on techniques applicable to linear models. The performance of the resulting algorithms is evaluated and compared in simulations, which show that our methods can, in many cases, provide useful information on causal relationships even for relatively small sample sizes.
\end{abstract}

\section{Introduction}
\label{sec:Introduction}

Many techniques have recently been developed for inferring causal relationships from data over a set of random variables \cite{SpiGlySch00,Pea09,ChiMee02,ShiHoyHyvKer06,ShiInaSogHyvKawWasHoyBol11,Hyv10,KawBolShiWas10,JanHoySch10,ZscJanZha11}. While most of this work has focused on uncovering connections among scalar random variables, in many actual cases each of the variables of interest may consist of multiple related, but distinct, measurements. For instance, in fMRI data analysis one is often interested in the functional connectivity among brain \emph{regions}, and for each such region-of-interest one has data measured from a set of multiple voxels. Typically, in these cases, some aggregate of each area is computed, after which the standard approaches are directly applicable. However, it can be shown that not only may information be lost when computing aggregates, but the outputs of such methods may not even be correct in the large sample limit.

A simple example illustrating one of the problems inherent with working with aggregates is the following. Consider three sets of variables with causal connections $\sX \rightarrow \sY \rightarrow \sZ$, i.e.\ the variables in $\sX$ may influence the variables in $\sY$, but not directly the variables in $\sZ$, and the variables in $\sY$ may influence the variables in $\sZ$. In this case, each variable $x \in \sX$ is independent of each $z \in \sZ$ \emph{conditional on the full set of mediating variables} $\sY$. However, when replacing the variables of each group with their respective mean value (the typical aggregate used), denoted by $\bar{x}$, $\bar{y}$, and $\bar{z}$, in general we obtain $\bar{x} \notPerp \bar{z} \; | \; \bar{y}$ \cite{SpiGlySch00,SchSpi08}. Thus, it is important to develop methods for causal discovery that exploit the full information available, as opposed to only aggregates of the data.

Towards this end, in this paper we extend two existing approaches \cite{ShiInaSogHyvKawWasHoyBol11,Hyv10} designed for causal discovery among scalar random variables to the case of random vectors (i.e.\ groups of variables), both exploiting any kind of non-Gaussianity present in the data. We also extend a recent method \cite{JanHoySch10} for inferring the causal relationship among \emph{two} arbitrarily distributed multi-dimensional variables to an arbitrary number of such variables. After describing the resulting algorithms, we evaluate and compare their performance in numerical simulations.

\section{Model and Problem Statement}
\label{sec:Model}

Throughout the paper, we will use the term `group' to denote a set of underlying variables all belonging to the same (multi-dimensional) random vector representing a single concept (e.g.\ one region in fMRI analysis). We use the term `variable' to represent a single scalar random variable belonging to one of the groups. 
Thus, for $g=1,\ldots,G$, let $\sX_g$ denote group $g$, and let the random vector $\bx_g = (x_1^{(g)}, \ldots, x_{n_g}^{(g)})^T$ collect the $n_g$ random variables belonging to group $g$. We assume that the \emph{groups} $\sX_g$ can be arranged in a causal order $K = (k_1, \ldots, k_G)$, such that the causal relationships among the groups can be represented by a directed acyclic graph. The data generating process is assumed to be a set of linear equations, given by
\begin{equation}
\label{eq:model_groups}
 \bx_{k_i} = \sum_{j < i} \: \bB_{k_i,k_j} \bx_{k_j} + \be_{k_i}, \;\; i=1,\ldots,G,
\end{equation}
with $\bB_{k_i,k_j}$ arbitrary (real) matrices of dimension $n_{k_i} \times n_{k_j}$, containing the direct effects from group $\sX_{k_j}$ to group $\sX_{k_i}$. The vectors of disturbance terms $\be_{k_i}$ are assumed to be zero mean, and mutually independent over groups, i.e. $\be_{k_i} \Perp \be_{k_j}, \: i\neq j$, but are allowed to be dependent within each group. If we arrange the groups in a causal order $K$ and define $\bx = (\bx_{k_1}, \ldots, \bx_{k_G})$ and $\be = (\be_{k_1}, \ldots, \be_{k_G})$, we can rewrite Equation~\eqref{eq:model_groups} in matrix form as $\bx = \bB \bx + \be$ with $\bB$ a lower block triangular matrix.
The model reduces to standard LiNGAM (Linear Non-Gaussian Acyclic Model, \cite{ShiHoyHyvKer06,ShiInaSogHyvKawWasHoyBol11}) when $\forall g: n_g = 1$ and all disturbances $\be$ are non-Gaussian. It also includes the model of \cite{Hyv10} when $G=2$, $n_1 = n_2 = 1$ and the disturbances are non-Gaussian. Finally, it contains as a special case the noisy model of \cite{JanHoySch10} when $G=2$ (but with no restriction on the $n_g$ and $\be$).

We assume that all variables in $\bx$ are observed, and that the grouping of these variables is known. Given merely observations of $\bx$ generated by Model~\eqref{eq:model_groups} (i.e. $\bB$ and $\be$ are unknown), we want to estimate the unknown causal order $K$ among these groups. We denote the data matrix of observations over the variables $\bx$ as $\bX = (\bX_1, \ldots, \bX_G)^T$, where each column corresponds to one observation and each row to one variable. The observations are grouped according to the $G$ groups, arranged in a random order, such that the first $n_1$ rows correspond to group $\sX_1$, the following $n_2$ rows to group $\sX_2$, and so on.

We note that our model family is equivalent to that given by \cite{KawBolShiWas10}. The main difference between our approach and theirs is that they do not assume to know which variable belongs to what group, which results in algorithms exponential in the number of involved variables, whereas our algorithm explicitly builds upon such knowledge, allowing to construct computationally and statistically more efficient algorithms, polynomial in the number of groups.

\section{Method}
\label{sec:alg}

\vspace{-1mm}
The overall algorithm for finding a causal order among the groups follows the approach introduced in \cite{ShiInaSogHyvKawWasHoyBol11}. We first search for an exogenous group (Section~\ref{sec:exogVar}), and then `regress out' the effect of this group on all other groups (Section~\ref{sec:LearnOrder}). We iterate this process to generate a full causal order over the $G$ groups.

\vspace{-1mm}
\subsection{Finding an Exogenous Group}
\label{sec:exogVar}

\vspace{-0.5mm}
We generalize three existing methods to search for an exogenous group, formally defined as below. Note that since the connections among the groups are assumed to be acyclic, there always exists at least one such exogenous (`source') group.

\vspace{-1mm}
\begin{mydef}
A group $\sX_j$ is \emph{exogenous} if for $\bx_j$ all matrices $\bB_{j,i}$ of Equation~\eqref{eq:model_groups} are zero.
\end{mydef}

\vspace{-4mm}
\subsubsection{GroupDirectLiNGAM}
\label{sec:GroupDirectLiNGAM}

As our first approach, we generalize the idea of DirectLiNGAM \cite{ShiInaSogHyvKawWasHoyBol11} to find an exogenous \emph{variable}, to finding an exogenous \emph{group}. The following lemma, which corresponds directly to Lemma~1 in \cite{ShiInaSogHyvKawWasHoyBol11}, states a criterion to find an exogenous group using regressions and independence tests.

\vspace{-1mm}
\begin{lemma}
\label{lemma:exogenous_LiNGAM}
Let $\bx$ follow Model~\eqref{eq:model_groups} with non-Gaussian disturbance terms $\be$. Let $\br_i^{(j)} := \bx_i - \bC \bx_j$ be the residuals when regressing $\bx_i$ on $\bx_j$ using ordinary least squares (OLS). A group $\sX_j$ is exogenous if and only if $\bx_j \Perp \br_i^{(j)}$ for all $i\neq j$.
\end{lemma}
The proof of this lemma, and the proof of Lemma~\ref{lemma:residuals} in Section~\ref{sec:LearnOrder} are left to the online appendix at \url{http://www.cs.helsinki.fi/u/entner/GroupCausalOrder/}

To apply Lemma~\ref{lemma:exogenous_LiNGAM} in practice, we need to test for (in)dependence between two vectors of random variables, and combine the results of several such independence tests. Assuming that the test returns p-values $p_{ji}$ under the null hypothesis of $\bx_j \Perp \br_i^{(j)}, \: i\neq j,$ we can get a measure of how exogenous group $\sX_j$ is by combining these p-values using Fisher's method \cite{Fis50}. This means that we select as the exogenous group the one minimizing
\begin{align}
\label{eq:logsumming}
\exm^{(j)} = - \sideset{}{_{i \neq j}}\sum \log(p_{ji}).
\end{align}

To obtain the p-values $p_{ji}$ we can test for joint dependence of the two vectors $\bx_j$ and $\br_i^{(j)}$ using the Hilbert Schmidt Independence Criterion (HSIC, \cite{GreFukTeoSonSchSmo08}), which, however, requires many samples to detect dependencies for high dimensional vectors. Alternatively, we can perform pairwise tests of each variable in $\bx_j$ against each variable in $\br_i^{(j)}$ using nonlinear correlations, and combine the resulting $n_j \times n_i$ p-values appropriately. Details are left to the online appendix.

\subsubsection{Pairwise Measure}
\label{sec:Pairwise}

Our second approach is based on modifying the pairwise measure \cite{Hyv10} designed for inferring the causal relationship between two linearly related non-Gaussian scalar random variables $x$ and $y$. If the true underlying \pagebreak
causal direction is from $x$ to $y$, the model is defined as $y = \rho x +e_y$ with $x \Perp e_y$. 
As pointed out in Section 2, this is just a special case of our more general model. 
The (normalized) ratio of the log likelihoods for the two possible causal models is given by 
$ R(x,y) = \left( \log  L(x\rightarrow y) - \log L(y\rightarrow x) \right) / \nsamp, $ 
where $\nsamp$ is the sample size and $L$ the likelihood of the specified direction, under some suitable assumption on the distributions of the disturbances. If the true underlying causal direction is $x\rightarrow y$, then $R(x,y) >0$ in the large sample limit. Symmetrically, if $x \leftarrow y$, then $R(x,y)<0$ in the limit. 

To use the ratio $R(\cdot \,, \cdot)$ to find an exogenous group $\sX_j$, the na\"{\i}ve approach is to calculate $R(x_k^{(j)}, x_l^{(i)})$ for each pair with $x_k^{(j)} \in \sX_j, \, k=1,\ldots,n_j, \, x_l^{(i)} \in \sX_i, \, l=1,\ldots,n_i, \, i\neq j$, and combine these measures. However, even if $\sX_j$ is exogenous, these pairs do not necessarily meet the model assumption because of the dependent error terms within each group, and hence there is no guarantee for correctness even in the large sample limit. This approach, termed the \textit{Na\"{\i}ve Pairwise Measure}, may however have a statistical advantage for small sample sizes (see Section~\ref{sec:experiments}).

To obtain a consistent method (simply termed \textit{Pairwise Measure} in Section~\ref{sec:experiments}), we replace the second variable of the pairs $(x_k^{(j)}, x_l^{(i)})$ with a quantity which guarantees that the model assumption is met if $\sX_j$ is exogenous: We first estimate the regression model $x_l^{(i)} = \sum_{\tilde{k}=1}^{n_j} \hat{b}_{l \tilde{k}} x_{\tilde{k}}^{(j)} + r_{l,(i)}$. If $\sX_j$ is exogenous then the regression coefficients $\hat{b}_{l \tilde{k}}$ are consistent estimators of the true total effects (when marginalizing out any intermediate groups). Hence, defining $z_{k,l}^{(i)} := x_l^{(i)} - \sum_{\substack{\tilde{k}=1;  \tilde{k} \neq k}}^{n_j} \hat{b}_{l \tilde{k}} x_{\tilde{k}}^{(j)} = \hat{b}_{lk} x_k^{(j)} + r_{l,(i)}$ yields a pair $(x_k^{(j)}, z_{k,l}^{(i)})$ meeting the model assumption of \cite{Hyv10} if $\sX_j$ is exogenous. Thus, in this case, $R(x_k^{(j)}, z_{k,l}^{(i)}) > 0$, in the limit, for all $k, l,$ and $i \neq j$. 
On the contrary, if $\sX_j$ is not exogenous the measure can take either sign, and simulations show that it is unlikely to always obtain a positive one. A way to combine the ratios is suggested in \cite{Hyv10}, which can be modified for the group case as
\begin{equation}
\label{eq:minsquare}
\exm^{(j)} = \frac{1}{n_j \sum_{i \neq j} n_i} \sum_{k=1}^{n_j} \sum_{i \neq j} \sum_{l=1}^{n_i} \min\{0, R(x_k^{(j)},z_{k,l}^{(i)})\}^2.
\end{equation}
That is, we penalize each negative value according to its squared magnitude and adjust for the group sizes. We select the group minimizing this measure as the exogenous one.

\vspace{-1mm}
\subsubsection{Trace Method}
\label{sec:TraceMethod}

Our third method for finding an exogenous group is based on the approach of \cite{JanHoySch10,ZscJanZha11}, termed the Trace Method, designed to infer the causal order among two groups of variables $\sX$ and $\sY$ with $n_x$ and $n_y$ variables, respectively. If the underlying true causality is given by $\sX \rightarrow \sY$, the model is defined as $\by = \bB \bx + \be$, where the connection matrix $\bB$ is chosen independently of the covariance matrix of the regressors $\bSigma := \cov(\bx, \bx)$, and the disturbances $\be$ are independent of $\bx$. Note that this method is based purely on second-order statistics and does not make any assumptions about the distribution of the error terms $\be$, as opposed to the previous two approaches where we needed non-Gaussianity. 
The measure to infer the causal direction defined in \cite{JanHoySch10} is given by

\begin{align}
\label{eq:delta}
   \Delta_{\sX \rightarrow \sY} := \log \left(  tr(\hat{\bB} \hat{\bSigma} \hat{\bB}^T) / n_y \right) - \log \left(  tr( \hat{\bSigma} ) / n_x \right) 
   - \log \left( tr(\hat{\bB} \hat{\bB}^T) / n_y \right)
\end{align}
where $tr(\cdot)$ denotes the trace of a matrix, $\hat{\bSigma}$ an estimate of the covariance matrix of $\bx$, and $\hat{\bB}$ the OLS estimate of the connection matrix from $\bx$ to $\by$. 
The measure for the backward direction $\Delta_{\sY \rightarrow \sX}$ is calculated similarly by exchanging $\hat{\bB}$ with the OLS estimate of the connection matrix from $\by$ to $\bx$ and $\hat{\bSigma}$ with the estimated covariance matrix of $\by$. If the correct direction is given by $\sX \rightarrow \sY$, Janzing et al. \cite{JanHoySch10} (i) conclude that $\Delta_{\sX \rightarrow \sY} \approx 0$, (ii) show for the special case of $\bB$ being an orthogonal matrix and the covariance matrix of $\be$ being $\lambda \bI$, that $\Delta_{\sY \rightarrow \sX} < 0$, and (iii) show for the noise free case that $\Delta_{\sY \rightarrow \sX} \geq 0$. Hence, the underlying direction is inferred to be the one yielding $\Delta$ closer to zero \cite{JanHoySch10}. In particular, if $|\Delta_{\sX \rightarrow \sY} |\, / \,| \Delta_{\sY \rightarrow \sX} | < 1$, then the direction is judged to be $\sX \rightarrow \sY$.

We suggest using the Trace Method to find an exogenous group $\sX_j$ among $G$ groups in the following way. For each $j$, we calculate the measures $\Delta_{\sX_j \rightarrow \sX_i}$ and $\Delta_{\sX_i \rightarrow \sX_j}$, for all $i \neq j$, and infer as exogenous group the one minimizing
\vspace{-1mm}
\begin{align}
\label{eq:ratiosquare}
\exm^{(j)} = \sideset{}{_{i \neq j}} \sum \left( \Delta_{\sX_j \rightarrow \sX_i} \, / \, \Delta_{\sX_i \rightarrow \sX_j} \right)^2.
\end{align}

\vspace{-3mm}
\subsection{Estimating a Causal Order}
\label{sec:LearnOrder}

Following the approach of \cite{ShiInaSogHyvKawWasHoyBol11}, after finding an exogenous group we `regress out' the effect of this group on all other groups. Since the resulting data set follows again the model in Equation~\eqref{eq:model_groups} having the same causal order as the original groups, we can search for the next group in the causal order in this reduced data set. This is formally stated in the following lemma, which corresponds to the combination of Lemma~2 and Corollary~1 in \cite{ShiInaSogHyvKawWasHoyBol11}.
\begin{lemma}
\label{lemma:residuals}
Let $\bx$ follow Model~\eqref{eq:model_groups}, and assume that the group $\sX_j$ is exogenous. Let $\br_i^{(j)} := \bx_i - \bC \bx_j$ be the residuals when regressing $\bx_i$ on $\bx_j$ using OLS, for $i=1,\ldots,G, \: i\neq j$, and denote by $\br^{(j)} $ the column vector concatenating all these residuals. Then $\br^{(j)} = \bB^{(j)} \br^{(j)} + \be^{(j)}$ follows Model~\eqref{eq:model_groups}. 
Furthermore, the residuals in $\br_i^{(j)}$ follow the same causal order as the original groups $\bx_i, \: i \neq j$. 
\end{lemma}
Using Lemma~\ref{lemma:residuals}, and the methods of Section~\ref{sec:exogVar}, we can formalize the approach to find a causal order among the groups as shown in Algorithm~\ref{alg:CausalRegionsGroups}.

\begin{algorithm}[tb]
   \caption{(Estimating a Causal Order among Groups)}
   \label{alg:CausalRegionsGroups}
\vspace{1mm}
\textbf{Input:} Data matrix $\bX$ generated by Model~\eqref{eq:model_groups}, arranged in a random causal order

\begin{algorithmic}
   \STATE Initialize the causal order $K:=[\,]$.
   \REPEAT
   \STATE Find an exogenous group $\sX_j$ from $\bX$ using one of the approaches in Section~\ref{sec:exogVar}.
   \STATE Append $j$ to $K$.
   \STATE Replace the data matrix $\bX$ with the matrix $\bR^{(j)}$ concatenating all residuals $\bR_{i}^{(j)}, \: i\neq j,$ from the regressions of $\bx_i$ on $\bx_j$ using OLS:
     \vspace{-2mm}
       \begin{equation*}
          \bX_{i}   = \bC_{i,j} \bX_{j} + \bR_i^{(j)} \; \text{ with } 
          \bC_{i,j} = \cov(\bX_i,\bX_j) \: \cov(\bX_j,\bX_j)^{-1}.
       \end{equation*}
     \vspace{-5mm}
   \UNTIL{$G-1$ group indices are appended to $K$}
   \STATE Append the remaining group index to $K$.
\end{algorithmic}
\end{algorithm}

\vspace{-1.5mm}
\subsection{Handling Large Variable Sets with Few Observations}
\label{sec:largeSets}

The OLS estimation used in Algorithm~\ref{alg:CausalRegionsGroups} requires an estimate of the inverse covariance matrix which can lead to unreliable results in the case of low sample size. 
One approach to solving this problem is to use regularization. 
For the $L^2$-regularized estimate of the connection matrix we obtain $\hat{\bC}_{i,j} = \bX_i \bX_j^T \: (\bX^T_j \bX_j + \lambda \bI)^{-1} \: = \cov(\bX_i,\bX_j) \: \nsamp$ $(\nsamp \, \cov(\bX_j,\bX_j) + \lambda \bI)^{-1}$, with $\nsamp$ the sample size and $\lambda$ the regularization parameter, see for example \cite{HasTibFri08}. 
In particular, this provides a regularized estimate of the covariance matrix.

Another approach is to apply the methods of Section~\ref{sec:exogVar} for finding an exogenous group to $N$ data sets, each of which consists of $G$ groups formed by taking subsets of the variables of the corresponding original groups. We then calculate measures $\exm^{(j)}_n, \, $ $j=1,\ldots,G, \: n=1,\ldots,N$, as in Equations~\eqref{eq:logsumming}, \eqref{eq:minsquare} or \eqref{eq:ratiosquare}, for each such data set separately, and pick the group $\sX_{j^*}$ which minimizes the sum over these sets to be an exogenous one, i.e.
\begin{equation}
\label{eq:combine_subgroups}
 j^* = \arg \min_j \: \sideset{}{_{1 \leq n \leq N}} \sum \exm^{(j)}_n
\end{equation}
where $\exm^{(j)}_n$ is the measure of group $j$ in the $n^{th}$ data set. We then can proceed as in Algorithm~\ref{alg:CausalRegionsGroups} to find the whole causal order.

Note that the same approach can be used when multiple data sets are available, which are assumed to have the same causal order among the groups but possibly different parameter values. An example for such a scenario is given by fMRI data from several individuals. An equivalent of Equation~\eqref{eq:combine_subgroups} was suggested in \cite{Shi12} for the single variable case with multiple data sets.

\section{Simulations}
\label{sec:experiments}

\begin{figure*}
\centering
\subfigure[100 models with 5 groups]{ 
\includegraphics[trim = 2mm 0mm 8mm 1mm, clip, width=.20\textwidth]
{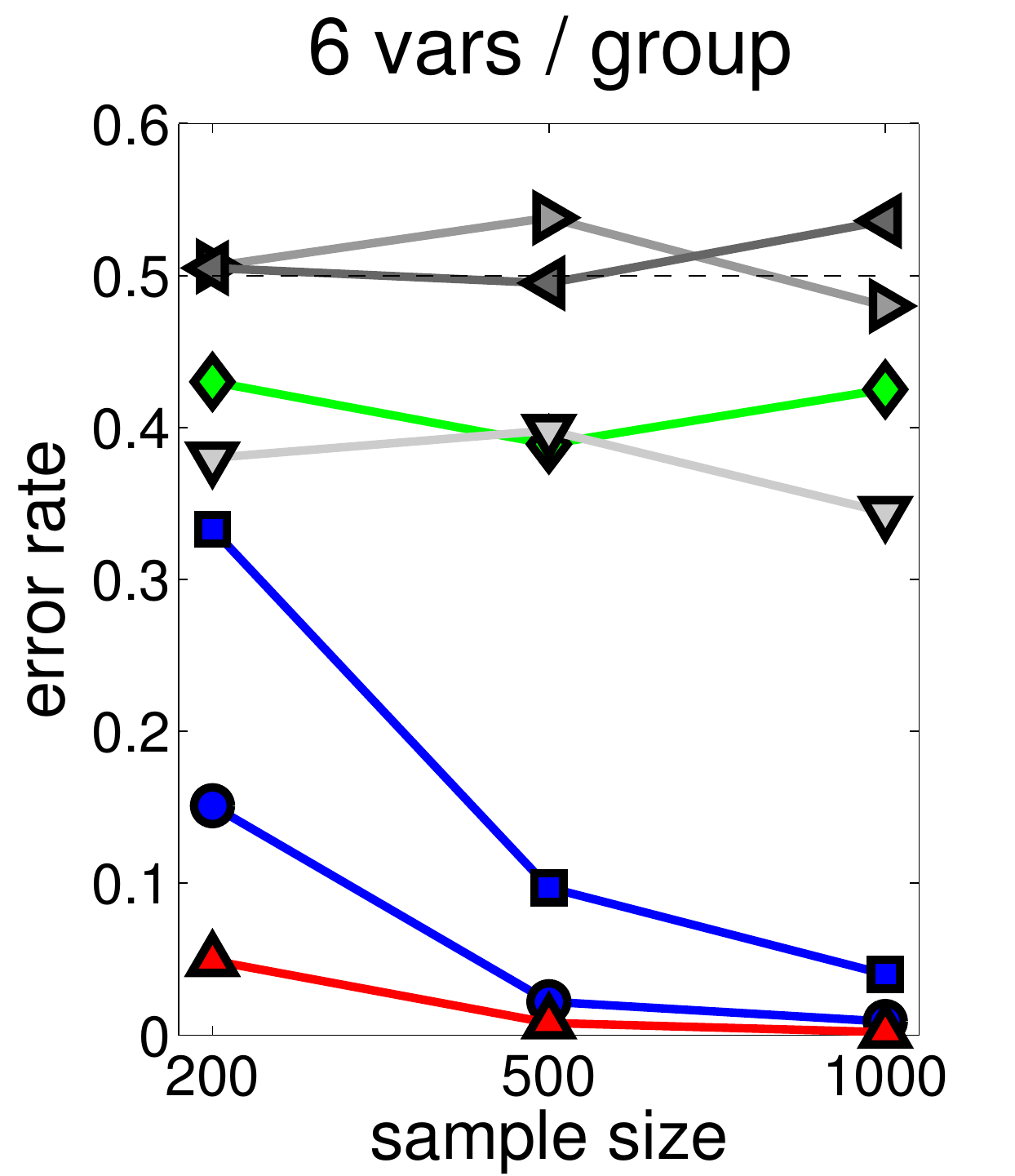} \hspace{-1mm}
\includegraphics[trim = 2mm 0mm 8mm 1mm, clip, width=.20\textwidth]
{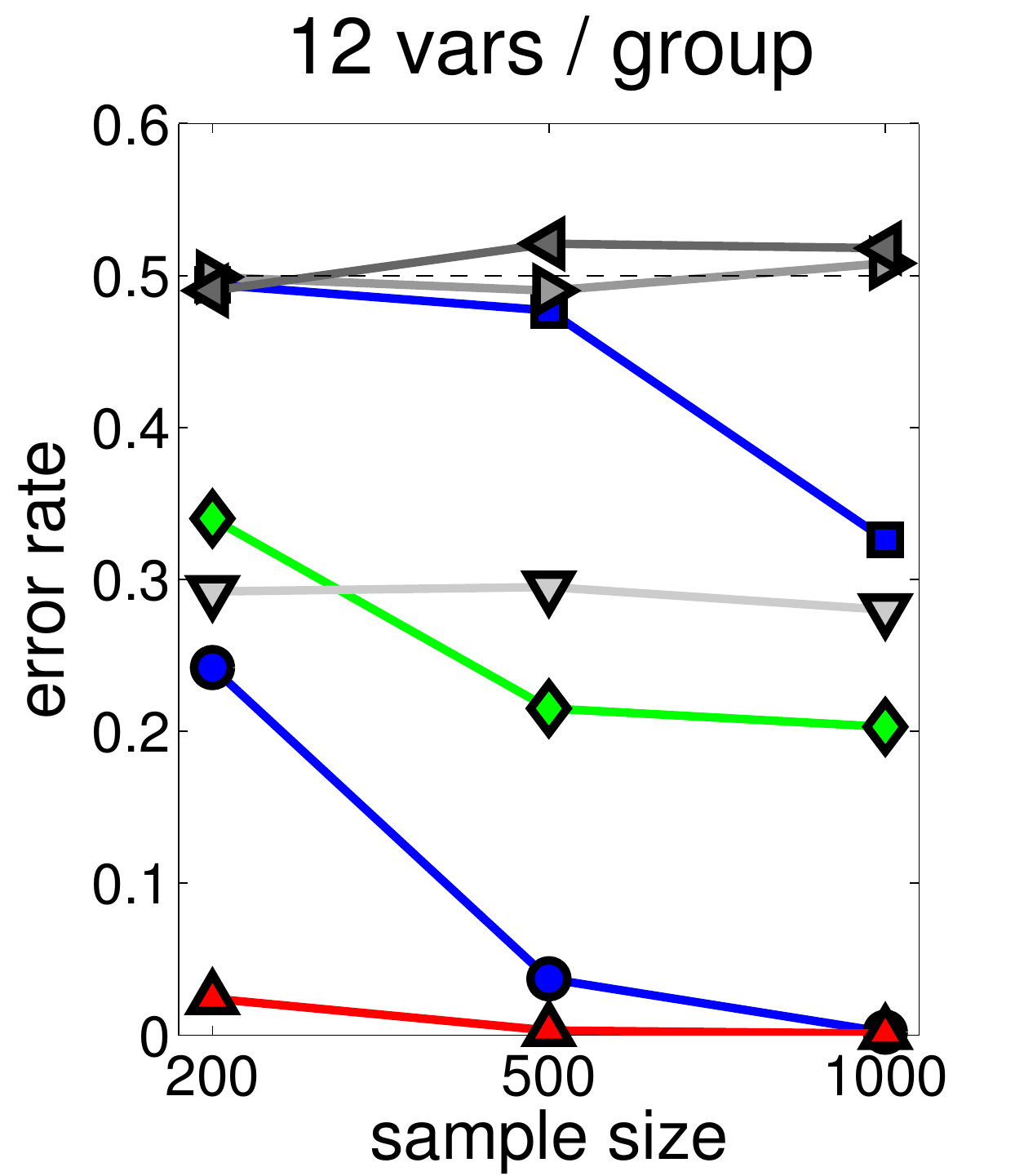} \hspace{-1mm}
\includegraphics[trim = 5mm 40mm 61mm 1mm, clip, width=.13\textwidth]
{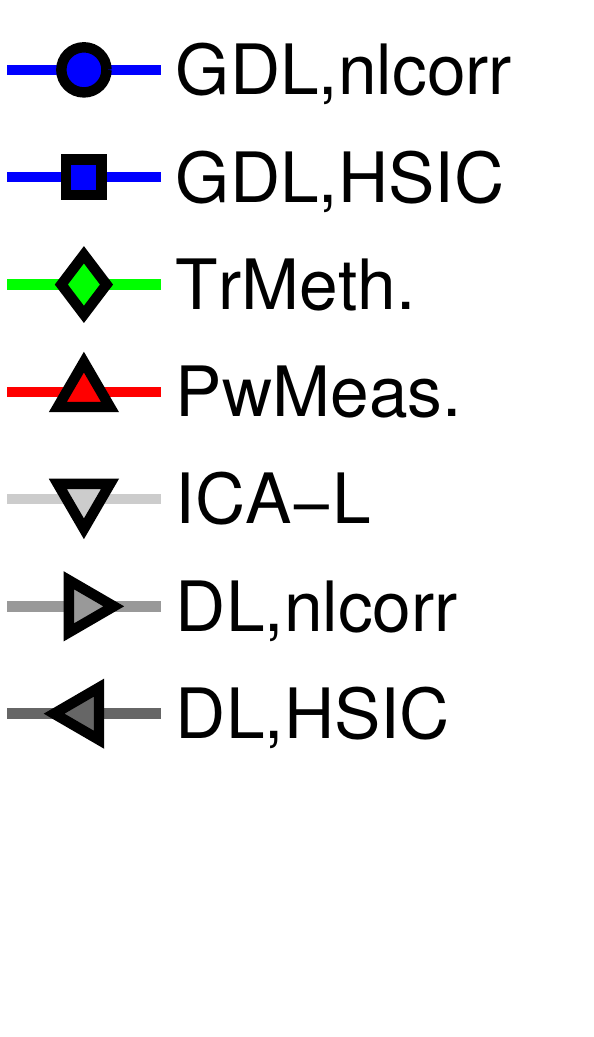}
}
\subfigure[50 models with 3 groups]{
\includegraphics[trim = 2mm 1mm 5mm 1mm, clip, width=.205\textwidth]
{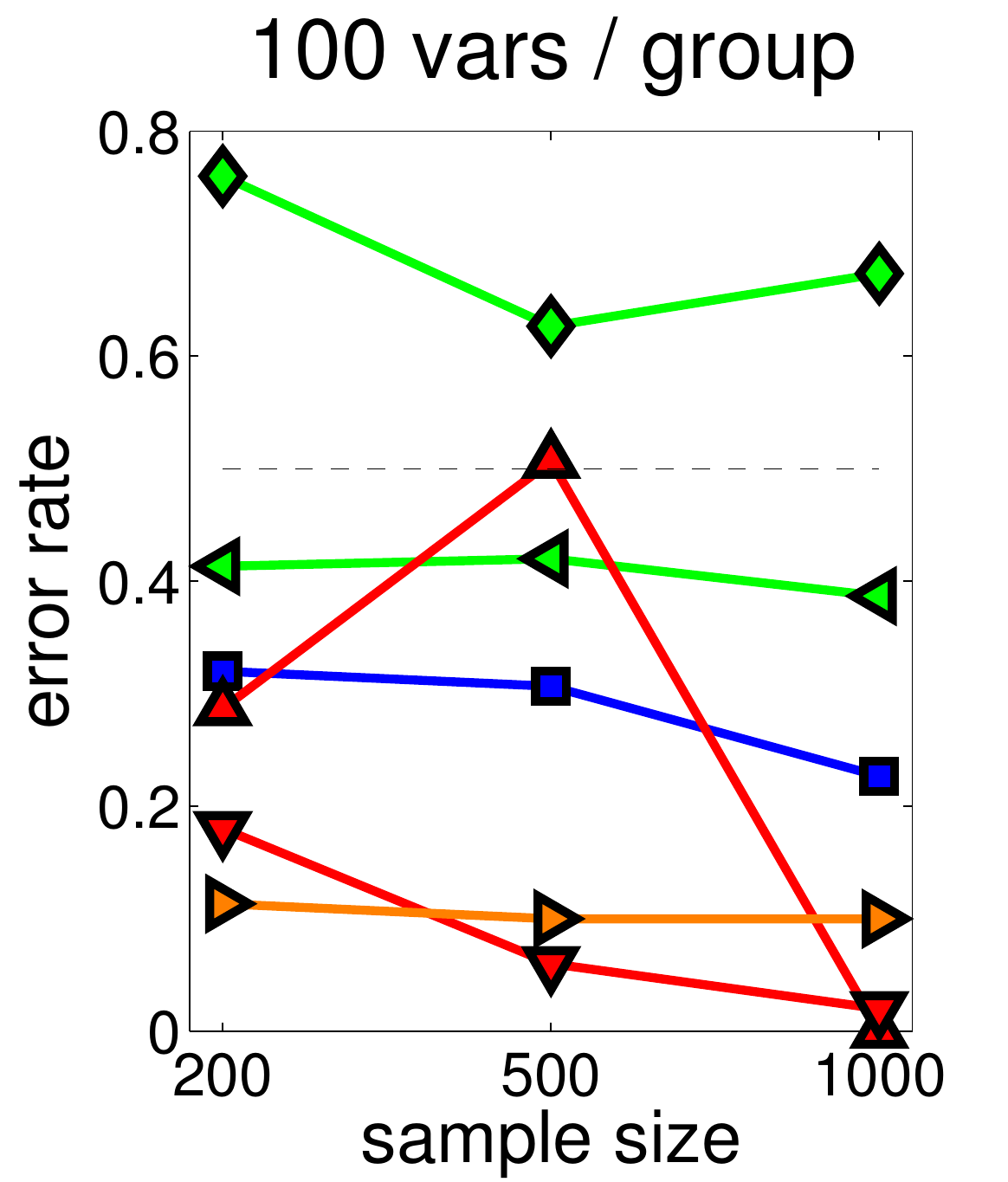} \hspace{-2mm}
\includegraphics[trim = 4mm 41.5mm 40mm 1mm, clip, width=.18\textwidth]
{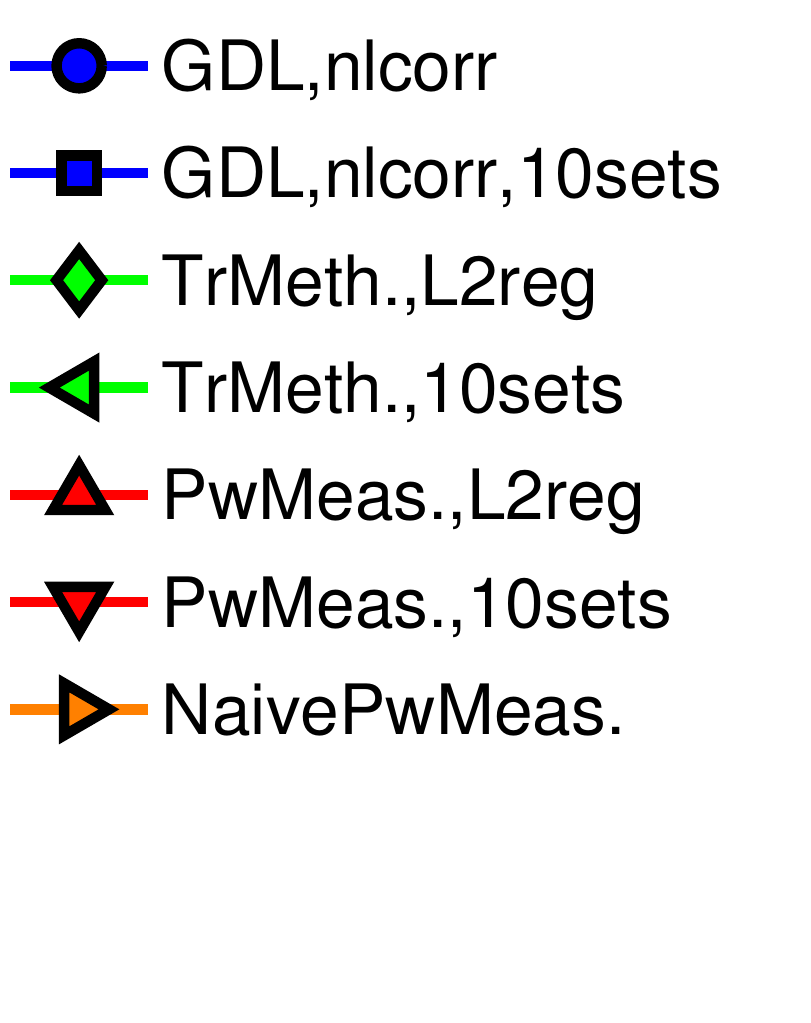}
}
\vspace{-1mm}
\caption{Sample size (x-axis) against error rate (y-axis) for various model sizes and algorithms, as indicated in the legends (abbreviations: GDL = GroupDirectLiNGAM; nlcorr, HSIC: nonlinear correlation or HSIC as independence test; TrMeth. = Trace Method; PwMeas. = Pairwise Measure; ICA-L = modified ICA-LiNGAM approach; DL = DirectLiNGAM on the mean-variables; 10sets = Equation~\eqref{eq:combine_subgroups} on $N=10$ data sets; L2reg =  $L^2$-regularization for covariance matrix). The dashed black line indicates the number of mistakes made when randomly guessing an order.}\vspace{-2mm}
 \label{fig:samp_conv}
\end{figure*}

Together, the methods of Section 3 provide a diverse toolbox for inferring the model of Section 2. Here, we provide simulations to evaluate the performance of the variants of Algorithm 1, and compare it to a few ad hoc methods. Matlab code is available at \url{http://www.cs.helsinki.fi/u/entner/GroupCausalOrder/}

We generate models following Equation~\eqref{eq:model_groups} by randomly creating the connection matrices $\bB_{k_i,k_j}, \: i > j$ with, on average, $s$\% of the entries being nonzero and additionally ensure that at least one entry is nonzero, to ensure a complete graph over the groups. To obtain the disturbance terms $\be_{k_i}$ for each group, we linearly mix random samples from various independent non-Gaussian variables as to obtain dependent error terms within each group. Finally, we generate the sample matrix $\bX$ and randomly block-permute the rows (groups) to hide the generating causal order from the inference algorithms.

We compare the variants of Algorithm~\ref{alg:CausalRegionsGroups} to two ad hoc methods. The first one is a modified ICA-based LiNGAM approach \cite{ShiHoyHyvKer06} where instead of searching for a permutation yielding a lower triangular connection matrix $\bB$ (i.e. finding a causal order among the \emph{variables}), we search for a block permutation yielding a lower block triangular matrix $\bB$ (i.e. finding a causal order among the \emph{groups}). Secondly, we compare our approach to DirectLiNGAM \cite{ShiInaSogHyvKawWasHoyBol11}, when replacing each group by the mean of all its variables.\footnote{We do not compare our results to methods such as PC \cite{SpiGlySch00} or GES \cite{ChiMee02}, as they cannot distinguish between Markov-equivalent graphs. Hence, in these simulations, they cannot provide any conclusions about the ordering among the groups since we generate complete graphs over the groups to ensure a total causal order.} 

We measure the performance of the methods by computing the error rates for predicting whether $\sX_i$ is prior to $\sX_j$, for all pairs $(i,j),\, i < j$.

Results for simulated data of sample size 200, 500 and 1000 generated from 100 random models having 5 groups with either $6$ or $12$ variables each, and $s=10\%$, are shown in Figure~\ref{fig:samp_conv}~\sfa.  
As expected, most methods based on Algorithm~\ref{alg:CausalRegionsGroups} 
improve their performance with increasing sample size. The only exception is the Trace Method on the smaller models; to be fair the method was not really designed for so few dimensions. Overall, the best performing method is the Pairwise Measure, closely followed by GroupDirectLiNGAM for the larger sample sizes. 
The ad hoc methods using DirectLiNGAM on the mean perform about as well as guessing an order (indicated by the dashed black line), whereas the modified ICA-LiNGAM approach performs better than guessing. However, it does not seem to converge for growing sample size, probably due to the dependent errors within each group, which is a violation of the ICA model assumption. 

We next replace each group by a subset of its variables of size $m = 1,\ldots,n_g$, and apply Algorithm~\ref{alg:CausalRegionsGroups} to these subgroups. As expected, the larger $m$ is, the less ordering mistakes are made. Details can be found in the online appendix.

Finally, we test the strategies described in Section~\ref{sec:largeSets} for handling low sample sizes in high dimensions on 50 models with 3 groups of 100 variables each, using 200, 500 and 1000 samples, and $s=5\%$. For $L^2$-regularization, we choose the pa-\linebreak rameter $\lambda$ using 10-fold cross validation on the covariance matrix. When taking subgroups, we use $N = 10$ data sets, and each subgroup containing ten variables. The error rates are shown in Figure~\ref{fig:samp_conv}~\sfb\ (we only show the $L^2$-regularized results if they were better than without regularization). 
Unreliable estimates of the covariance matrix seem to affect especially the Trace Method, and the Pairwise Measure on the smaller sample sizes. On the smallest sample, using subsets seems to be advantageous for most methods, however, the best performing approach is the Na\"{\i}ve Pairwise Measure, which, however, does not seem to converge to be consistent, where as GroupDirectLiNGAM and the Pairwise Measure are.

In general, the simulations show that the introduced method often correctly identifies the true causal order, and clearly outperforms the simple ad hoc approaches. It is left to future work to study the performance in cases of model violations as well as to apply the method to real world data.

\vspace{-2.5mm}
\subsubsection*{Acknowledgments} We thank Ali Bahramisharif and Aapo Hyv\"{a}rinen for discussion. The authors were supported by Academy of Finland project \#1255625.

\vspace{-3mm}
\bibliographystyle{unsrt}
\bibliography{bibfile_abbreviated}

\end{document}